%% file: main.tex
\documentclass[runningheads]{llncs}
\usepackage{multirow}
\usepackage[table,xcdraw]{xcolor}
\usepackage[T1]{fontenc}
\usepackage{graphicx}
\usepackage{float}
\usepackage{url}
\usepackage{amssymb}
\usepackage{amsmath}
\usepackage{booktabs}
\usepackage{hyperref}
\begin{document}
\title{Graph Neural Networks: \\A suitable Alternative to MLPs in Latent \\3D Medical Image Classification?}
\titlerunning{Graph Neural Networks as an alternative to MLP's?}

\author{
Johannes Kiechle\inst{1,2,3} \and
Daniel M. Lang\inst{1,2} \and
Stefan M. Fischer\inst{1,2,3} \and \\
Lina Felsner\inst{1,2} \and
Jan C. Peeken\inst{1,2} \and
Julia A. Schnabel\inst{1,2,3,4}
}
\authorrunning{Kiechle et al.}

\institute{
Technical University of Munich, Germany \and
Helmholtz Munich, Germany \and
Munich Center for Machine Learning, Germany \and
King's College London, United Kingdom}

\maketitle

\begin{abstract}
Recent studies have underscored the capabilities of natural imaging foundation models to serve as powerful feature extractors, even in a zero-shot setting for medical imaging data. Most commonly, a shallow multi-layer perceptron (MLP) is appended to the feature extractor to facilitate end-to-end learning and downstream prediction tasks such as classification, thus representing the \textit{de facto} standard. However, as graph neural networks (GNNs) have become a practicable choice for various tasks in medical research in the recent past, we direct attention to the question of how effective GNNs are compared to MLP prediction heads for the task of 3D medical image classification, proposing them as a potential alternative. In our experiments, we devise a subject-level graph for each volumetric dataset instance. Therein latent representations of all slices in the volume, encoded through a DINOv2 pretrained vision transformer (ViT), constitute the nodes and their respective node features. We use public datasets to compare the classification heads numerically and evaluate various graph construction and graph convolution methods in our experiments. Our findings show enhancements of the GNN in classification performance and substantial improvements in runtime compared to an MLP prediction head. Additional robustness evaluations further validate the promising performance of the GNN, promoting them as a suitable alternative to traditional MLP classification heads. Our code is publicly available at: \url{https://github.com/compai-lab/2024-miccai-grail-kiechle}
\end{abstract}

\section{Introduction}

Deep learning-based methods in medical imaging research hold great promise for their ability to identify abnormalities or malicious diversification processes. These capabilities encompass early disease detection, diagnosis, characterization, or malignancy prediction, thereby enabling personalized therapies and potentially enhancing treatment efficacy and patient outcomes. Driven by technological advancements that have led to a wealth of tomography data, image analysis methods have gained increasing relevance in medical research. To this end, automated and effective extraction of imaging biomarkers from large-scale 3D medical imaging datasets constitutes the core component that substantially influences the efficacy of subsequent downstream tasks. \newline
\indent Foundation models have attracted significant attention in deep learning-based computer vision applications~\cite{kirillov2023segment,radford2021learning,zou2024segment}. They do not only show great promise for their ability to extract semantically rich features from imaging data but also for their easy adaptability to a wide range of downstream tasks such as segmentation, classification, or registration ~\cite{pai2024foundation,song2024general,wu2023medical}. While applying these powerful models has become standard practice in natural imaging, the progress in adapting them to medical imaging is proceeding more slowly. This is primarily due to the limited access to public large-scale medical imaging datasets required for training~\cite{azad2023foundational}. Nevertheless, recent studies have underscored the capability of vision foundation models, trained on natural images, being successfully transferred to medical imaging data~\cite{huix2024natural}. Typically, a shallow standard multi-layer perceptron (MLP) classifier is placed on top of the pretrained imaging encoder to enable joint end-to-end learning and facilitate downstream prediction tasks such as classification. The work by Truong et al.~\cite{truong2021transferable} uses MLP finetuning of multiple self-supervised ImageNet models for different medical detection and classification tasks using lymph node, fundus, or chest X-ray images. Nielsen et al.~\cite{nielsen2023self} also follow this approach, showing that state-of-the-art classification performance with only $1\%-10\%$ of the available labeled data can be achieved for three common medical imaging modalities (bone marrow microscopy, gastrointestinal endoscopy, dermoscopy). Baharoon et al.~\cite{baharoon2023towards} investigated the approach for radiology image analysis. They conducted 100 experiments across diverse modalities (X-ray, CT, and MRI) for classification and organ segmentation tasks, based on 2D and 3D images. \newline 
\indent With the emergence of graph neural networks (GNNs) ~\cite{scarselli2008graph}, effective modeling of naturally graph-structured data has become possible. GNNs are powerful at handling problems involving pairwise interactions between entities or modeling complex relational data by encoding relationships in the graph structure ~\cite{wu2020comprehensive}. This has led to successful applications in various domains, such as modeling drug interactions~\cite{zitnik2018modeling}, recommendation systems~\cite{ying2018graph}, population graphs~\cite{stankeviciute2020population,gu2024novel} or (bio)chemical molecules~\cite{gilmer2017neural,li2021structure}. Moreover, in the context of 3D data, GNNs have demonstrated an inherent ability to preserve the underlying spatial structure when the graph modeling is based on 2D representations~\cite{wei2020view}. This capability is likely to extend to 3D medical data, where GNNs can effectively represent spatial relationships between encoded slices within the input volume. Consequently, we posit that GNNs are better suited for being finetuned ontop of vision foundation models, such as DINOv2, using 3D medical data compared to MLPs. We argue that modeling the spatial representation of the input volume using graph-based latent encodings is crucial for understanding the 3D context of volumetric data, thus positioning GNNs as a powerful and legitimate alternative to MLPs.
\newline 
\indent In this work, we direct attention to the question of how effective GNNs are compared to MLP prediction heads for the task of latent 3D medical image classification based on the MedMNIST3D datasets~\cite{yang2023medmnist}. We encode input volumes according to their axial, sagittal, and coronal views and finetune GNN and MLP heads on top of a 2D DINOv2 pretrained ViT. To this end, we encode the topological inductive bias in the underlying graph structure to allow the GNN to learn based on the topology of the problem~\cite{xu2018howpowerful}.\newline
\indent Our main contributions and findings are as follows: \textbf{(I)} To the best of our knowledge, this work is the first to integrate a general vision foundation model with GNNs for the task of 3D medical image classification. \textbf{(II)} We demonstrate that GNNs, based on suitable graph topologies and convolution operations, can surpass the performance of MLPs in 3D medical image classification and maintain prediction performance on par with MLPs in robustness evaluations. \textbf{(III)} Our experiments indicate a substantial improvement in runtime for GNNs compared to MLPs, implying increased efficiency and a reduced carbon footprint. 

\begin{figure}[!t]
\centering
\includegraphics[width=\textwidth]{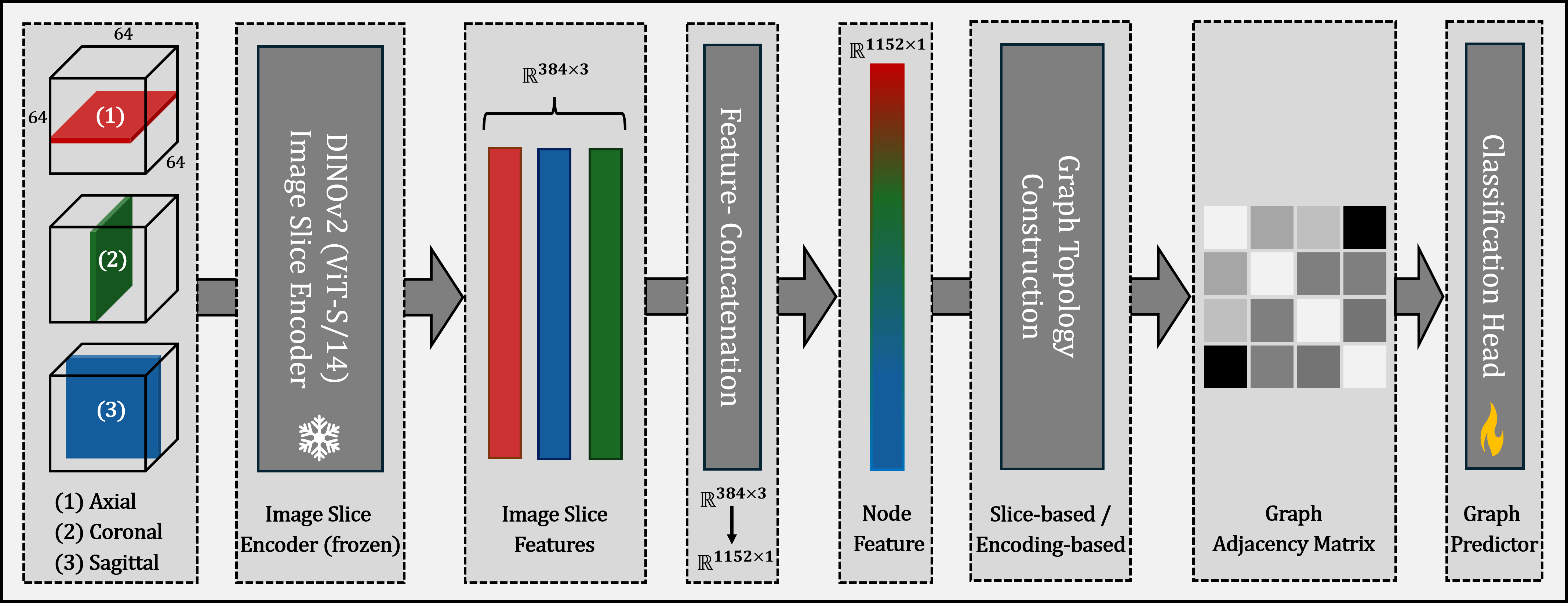}
\caption{Schematic block diagram of the proposed volumetric classification pipeline.} 
\label{fig:block_diagram}
\end{figure}

\section{Methods and Materials} \label{sec:methods}

Our proposed volumetric classification pipeline, as illustrated in \autoref{fig:block_diagram}, comprises the following steps: First, we encode 3D input volumes using all 64 axial, coronal, and sagittal views via a DINOv2 pretrained ViT. This encoding results in a $\mathbb{R}^{384 \times 3}$ feature representation per slice level, which we subsequently concatenate into a $\mathbb{R}^{1152 \times 1}$ feature vector. Thus, each volume (i.e., subject) is represented as $64$ feature vectors of size $1152$, which then undergo a graph topology construction step (see \autoref{subsec:graph_construction}) prior to the GNN-based classification.

\subsection{Volumetric Feature Encoding}

In this study, we employ DINOv2~\cite{oquab2023dinov2} as the image encoder. The original DINO (self-DIstillation with NO labels) model~\cite{caron2021emerging} is a self-supervised learning framework that employs a student-teacher architecture to extract semantically rich visual features from imaging data. Oquab et al.~\cite{oquab2023dinov2} introduced DINOv2, the current state-of-the-art network, which leverages self-supervised contrastive training on a large-scale dataset comprising 142 million images from curated and uncurated data sources. This results in representations that effectively capture the semantic meaning of images. DINOv2 has shown strong performance across various tasks, including pixel-level tasks, such as segmentation or detection, and image-level classification. In this work, we utilize a DINOv2 pretrained ViT model with $21$M parameters, which is frozen in our analysis and uses $14 \times 14$ patches to encode $224 \times 224$ axial, coronal, and sagittal input volume slices, resulting in a $384$-dimensional feature representation for each view. Input slices are linearly interpolated to match the desired input size.  

\subsection{Graph Representation Learning}


Let $G = (V,E)$ denote a graph with node attributes $X_v$ for $v \in V$ and edge attributes $e_{u,v}$ for $(u, v) \in E$. Given a set of graphs $\{G_N\}$ and their labels $\{y_N\}$, the task of graph supervised learning is to learn a representation vector $h_G \in \mathbb{R}^n$ that is used to predict the class of the entire graph, $y_G = g(h_G)$. To this end, GNNs update a node $v$'s feature representation by a non-linear function and permutation invariant aggregation of feature information from the neighborhood of $k$ nodes adjacent to the node $v$. In this study, we employ two graph convolution operator methods, namely Graph SAGE~\cite{hamilton2017inductive} and Graph Attention (GAT)~\cite{velivckovic2017graph}. 
To finally obtain a graph-level representation 
we apply a global, feature dimension-wise mean pooling operation. 

\subsection{Graph Construction} \label{subsec:graph_construction}
\begin{figure}[!t]
\centering
\includegraphics[width=0.90\textwidth]{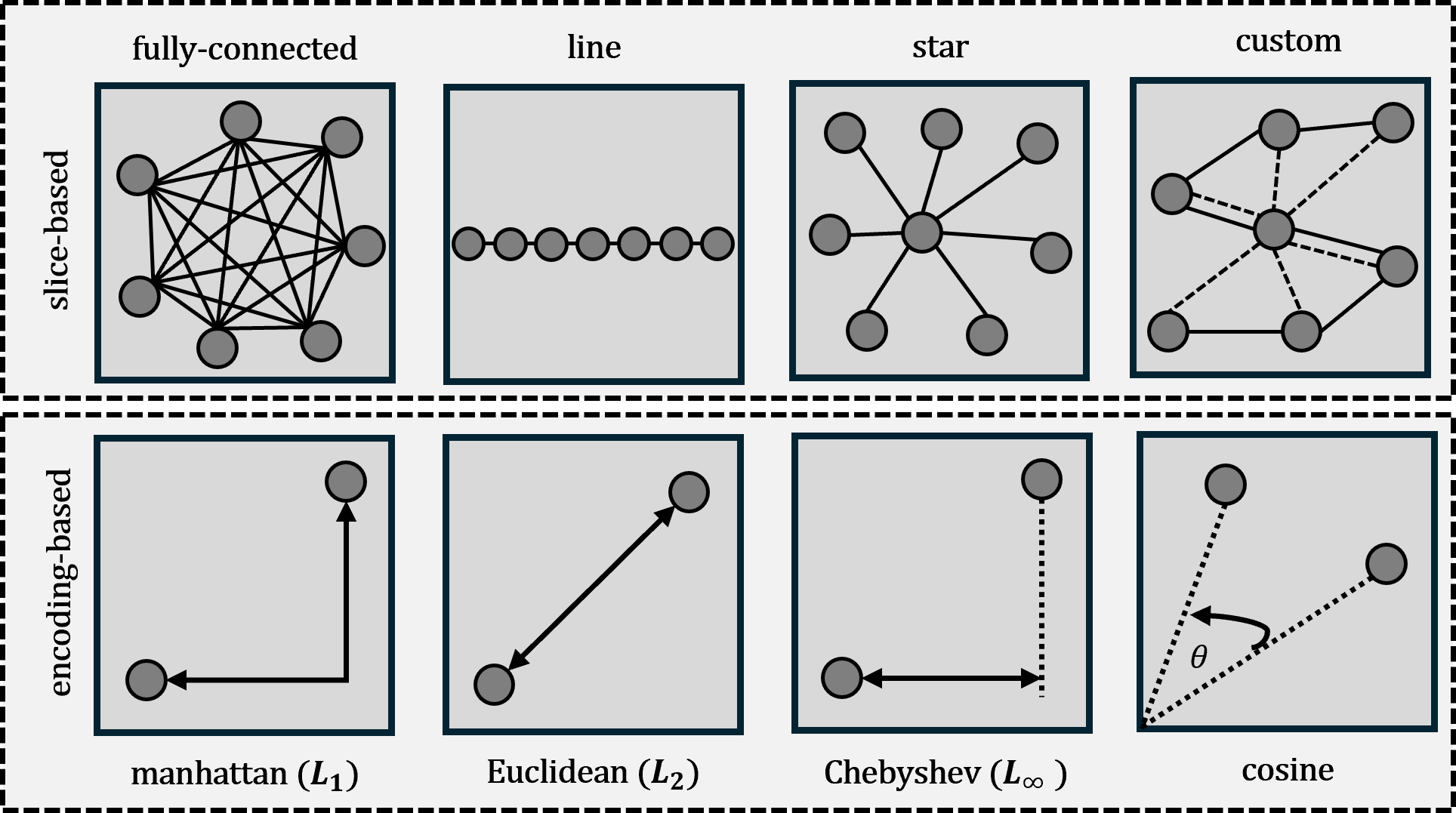}
\caption{Outline of the graph topologies which are used to evaluate the GNN downstream classification. The first row shows slice-based graph topology construction methods, whereas the second row represents the graph topologies based on the latent encodings.}
\label{fig:graph_topologies}
\end{figure}

We define the total number of nodes for each subject-level graph as the number of possible 2D views in the input volume. As all input images are isotropic in size of $64$ slices in each plane, the resulting graph does also comprise of 64 nodes. The final feature vector which describes a single node, is the concatenation of the DINOv2 encoded axial, coronal and sagittal slices according to their slice index. 
We define undirected graphs $G := (V,E)$ as a set of $|V|$ nodes and $|E|$ edges, where the graph topology is described by a symmetric adjacency matrix $A \in \mathbb{R}^{|V| \times |V|}$, where an entry $A_{u,v} = 1$ if $e_{uv} \in E$. For all our GNN experiments, we define the set of edges $|E|$ according to eight different graph construction strategies (\autoref{fig:graph_topologies}). We differentiate between \textit{slice-based} and \textit{encoding-based} graph topologies. \newline
\indent The first family of graph construction methods, denoted as \textit{slice-based topologies}, is based on the ordering (i.e., relative position) of the slices within the input volumes. We employ four construction strategies: fully connected, star, line, and custom graph topology (\autoref{fig:graph_topologies}, top). 
In the fully connected topology, every node is connected to every other node. The line topology maintains the sequential order of the slices as they appear in the input volume. The star and custom topologies connect the node representing the central slice to all other nodes. The custom topology extends the star topology by adding edge connections between neighboring nodes (i.e. consecutive slice indices). Since all samples in the datasets are centered around the lesion, we assume the central slices contain the most significant lesion area, thus enhancing useful information flow. \newline
\indent The second family of graph construction methods, denoted as \textit{encoding-based topologies}, is based on the node features (i.e., latent representations of the volume slices). We employ four different strategies for this construction (\autoref{fig:graph_topologies}, bottom). Specifically, we use three popular Minkowski distance metrics ($L_1$, $L_2$, and $L_\infty$) and cosine similarity. Given that the DINOv2 framework predominantly utilizes natural images for self-supervised training, our goal is to achieve a potentially better topological fit using encoding-based methods. This approach aims to facilitate zero-shot behavior when applied to medical imaging data.

\section{Experiments and Results}

We compare the classification performance of a GNN to that of a conditional MLP, both evaluated on MedMNIST3D datasets~\cite{yang2023medmnist}. To ensure a fair comparison and facilitate spatial awareness, the conditional MLP variable is represented by the respective slice number, appended to the $1152$-dimensional feature vector as illustrated in \autoref{fig:block_diagram}. This approach aligns with the graph topology construction methods, which are also based on the index of the volume slices. 

\subsection{Implementation and Training Details}
The DINOv2 image encoder is followed by a two-layer MLP or GNN classification head with interleaved ReLU activation. Following the MLP's best practices and to mitigate graph oversmoothing~\cite{keriven2022not}, we also maintain the GNN's depth at a shallow level. Both heads contain approximately 300k trainable parameters, ensuring a fair comparison. We use the official data splits for training, validation, and testing provided by the MedMNIST3D initiative~\cite{yang2023medmnist}. Training is conducted using the SGD optimizer with a weight decay of $0.1$, a batch size of $16$, an initial learning rate of $0.001$, and $300$ epochs. The GNN and MLP model achieving the best AUROC score on the validation set is selected for final evaluation on the test set. A learning rate scheduler decays the learning rate by $0.995$ each epoch to ensure smooth convergence. All models were trained on an Nvidia RTX A6000.

\begin{table}[!t]
\caption{Quantitative results. ACC represents the accuracy, AUCROC denotes the area under the receiver-operator characteristic curve. Runtime is given in minutes. We compare the baseline model from~\cite{yang2023medmnist} against conditional MLP and GNN prediction heads. The right column gives information about the GNN Configuration in terms of graph convolution operator (first entry) and graph topology (second entry), with $k$ as the number of neighbors per node. For reference, $\uparrow$ means that higher values are better. Results are reported as the mean over 3 runs with different random seeds and their standard deviations. For the baselines, only the mean is reported following \cite{yang2023medmnist}.}
\input{images/quantitative_results}
\label{tab1}
\end{table}
\begin{figure}[!t]
\centering
\includegraphics[width=\textwidth]{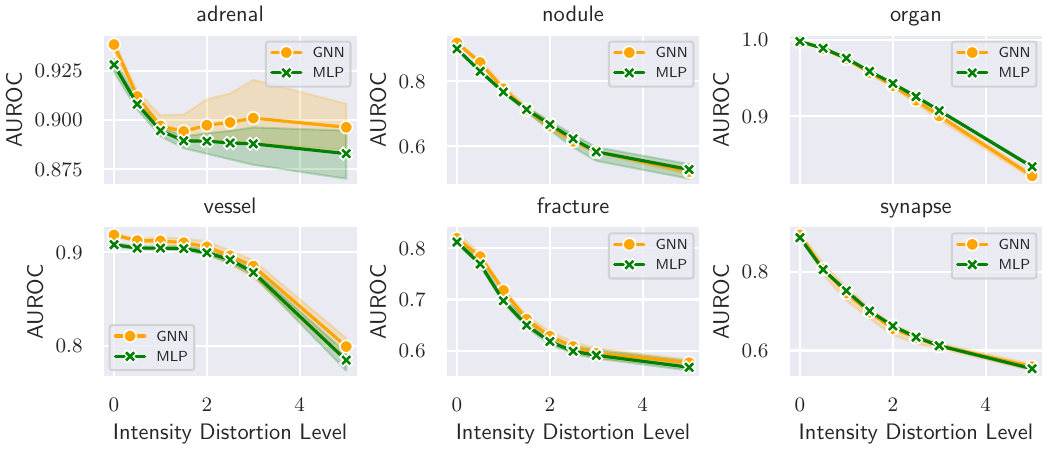}
\caption{Robustness evaluation of the best performing GNN versus the conditional MLP prediciton head. The x-axis represents the level of voxel intensity perturbation added to the initial 3D volume, whereas the y-axis shows the performance as the area under the receiver-operator characteristics (AUROC).} 
\label{fig:robustness}
\end{figure}

\subsection{Quantitative Results}
The results of our quantitative analysis are presented in~\autoref{tab1}. We evaluate the performance of all models using the area under the receiver-operating characteristics curve (AUROC) and accuracy (ACC) following~\cite{yang2023medmnist} to ensure fair comparison. Results are reported as the mean and standard deviation over three runs with different random seeds. As a baseline, we include the best-performing model for each MedMNIST3D dataset by Yang et al.~\cite{yang2023medmnist}. We report the total training time in minutes for the MLP and GNN models, excluding the MedMNIST3D baseline, which is trained from scratch, resulting in an unfair comparison. \newline
\indent At the core of our analysis, we evaluate our proposed GNN method against the MLP prediction head. We systematically conduct GNN experiments using various combinations of graph construction methods and graph convolution operations, as detailed in~\autoref{sec:methods}. \autoref{tab1} presents the best-performing GNN configuration and its corresponding metrics. Regarding AUROC, the GNN outperforms the MLP on all datasets except for OrganMNIST3D, where performance is equal. In terms of accuracy (ACC), the GNN also surpasses the MLP with minor improvements in four out of six use cases. \newline
\indent Notably, the stability with respect to topology varies across different datasets. The GNN topology appears to play a more substantial role for datasets with elongated or small structures, such as VesselMNIST3D and NoduleMNIST3D. Additionally, the results suggest that SAGEConv is particularly advantageous for slice-based topologies. Detailed AUROC results for all GNN configurations are provided in the supplementary material. \newline 
\indent The quantitative evaluation outlined in~\autoref{tab1}, shows that the MLP approach is able to outperform the baseline model in four out of six cases. Our proposed GNN method performs superior on all evaluated datasets for both AUROC and ACC, except for the case of VesselMNIST3D. This shows the good capabilities of GNNs and underlying power of DINOv2, which provide semantically meaningful features that prove useful for medical image classification. \newline
\indent We also report the runtime of the MLP and GNN prediction heads in~\autoref{tab1}. The GNN is substantially more efficient regarding computational cost than the MLP approach, with an average of only $41.7\%$ of the total runtime required by the MLP. This may not only benefit the overall efficacy but also reduce the carbon footprint, particularly in large-scale data training scenarios.

\subsection{Robustness Evaluation}
%
In addition to comparing the overall performance of the GNN and MLP prediction heads, we evaluate the robustness by simulating voxel intensity perturbations following~\cite{sudre2017longitudinal} to mimic various data sources and scanners. \autoref{fig:robustness} shows the AUROC values for different intensity distortion levels across all datasets of MedMNIST3D. These results indicates that the GNN maintains performance comparable to the MLP across various perturbation strengths. Except for the case of \mbox{AdrenalMNIST3D}, both models exhibit similar performance across increased distortion levels. For the AdrenalMNIST3D dataset, the GNN demonstrates enhanced robustness compared to the MLP, as shown by superior performance across increasing intensity distortions in the top-left plot of \autoref{fig:robustness}.


\section{Discussion and Conclusion}

In this paper, we evaluated the effectiveness of GNNs in comparison to MLPs, the \textit{de facto} standard for latent 3D medical image classification. We hypothesized that GNNs may better represent the spatial relationship of the encoded slices relative to their initial positioning in the input volume compared to MLPs. This hypothesis was based on the encoding of the input volume according to their axial, saggital and coronal views, using a 2D DINVOv2 pretrained ViT. 

Our results demonstrate that GNN classification heads can outperform traditional MLP prediction heads across various MedMNIST3D datasets, given a suitable GNN configuration. In addition to their enhanced predictive performance, our findings indicate that GNNs match MLPs in robustness analysis. Notably, our GNN experiments demonstrate a substantial reduction in runtime, almost less than $60\%$ on average compared to an MLP, despite an equivalent number of trainable parameters. Given that deep learning applications are highly energy-demanding, shorter runtimes will not only benefit efficiency but may also imply mitigating environmental impact by lowering carbon footprint. Based on the promising performance levels observed, promoting GNNs as a suitable alternative to traditional MLP classification heads is indeed a legitimate choice. 

Interestingly, our findings reveal that the optimal graph convolution operator and topology vary across different datasets. This necessitates identifying the most suitable graph convolution operator and underlying graph topology for each specific task, which is a notable limitation of the presented GNN approach. These results are consistent with existing studies indicating that the predictive efficacy of GNNs can be substantially influenced by the underlying graph structure~\cite{muller2023survey}. To address the challenge of topology identification, we plan to extend our work beyond static graph construction by incorporating adaptive methods~\cite{cosmo2020latent,kazi2022differentiable}. These methods, which learn the optimal graph structure towards the respective downstream task, are capable to show improved overall performance compared to static methods~\cite{muller2023survey}. In this context, we also intend to evaluate the proposed GNN classification method on additional datasets and test the GNN for other clinically relevant downstream tasks, such as prognosis. 





\begin{credits}
\subsubsection{\ackname} Author JK gratefully acknowledges financial funding received by the DAAD program Konrad Zuse Schools of Excellence in Reliable Artificial Intelligence, sponsored by the German Federal Ministry of Education and Research. Author JCP gratefully acknowledges funding from the Wilhelm Sander Foundation in Cancer Research (2022.032.1). Author SMF has received funding from the Deutsche Forschungsgemeinschaft (DFG, German Research Foundation) – 515279324 / SPP 2177.
\end{credits}

%
%
%

\bibliographystyle{splncs04}
\bibliography{mybibliography}





\end{document}


\title{Graph Neural Networks: \\A suitable alternative to MLPs in latent \\3D Medical Image Classification? \\Supplementary}
\titlerunning{Graph Neural Networks as an alternative to MLP's?}

\author{}

\institute{}


\begin{figure}[!b]
\centering
\includegraphics[width=\textwidth]{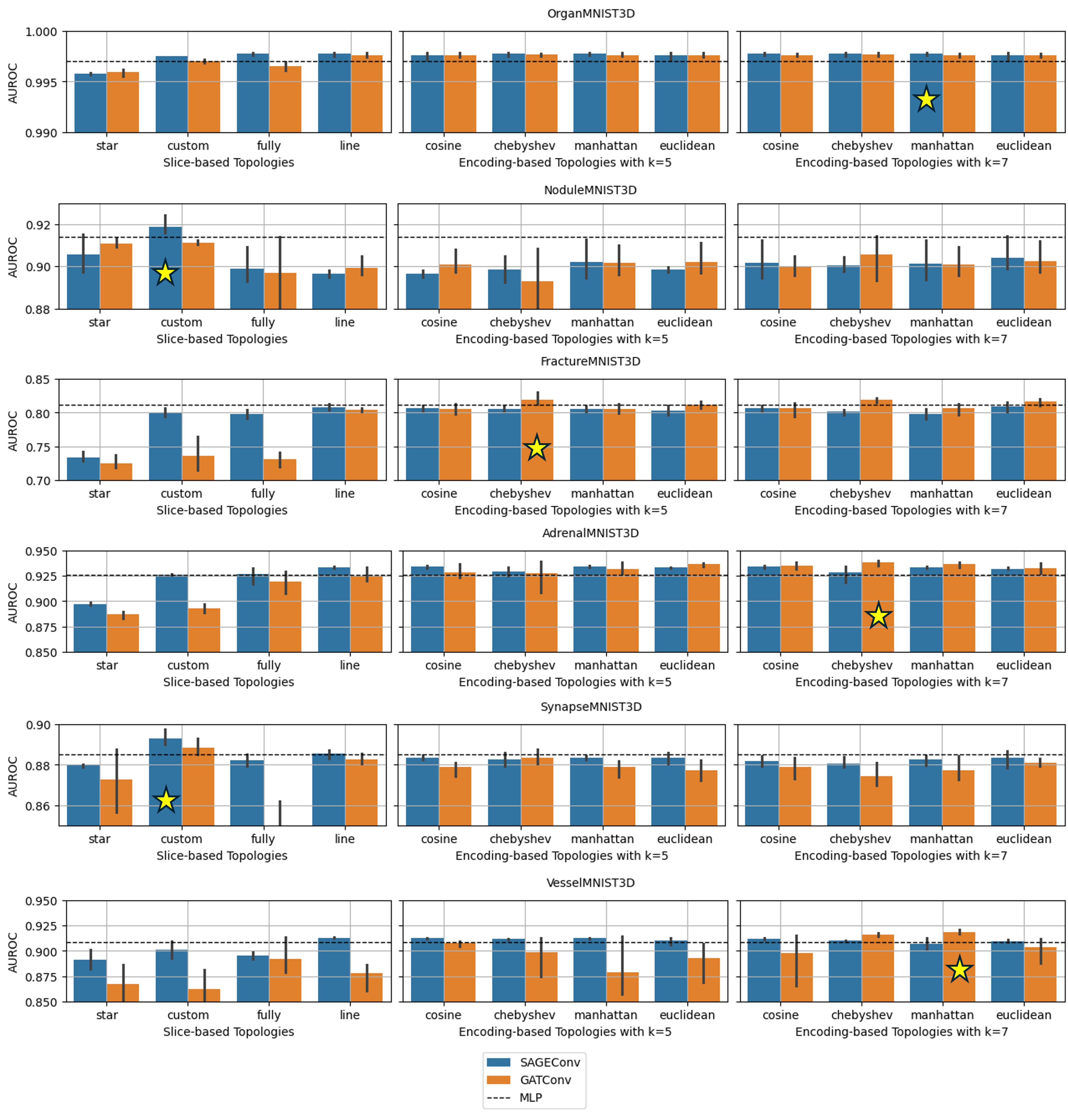}
\caption{Quantitative performance comparison of the GNN classification head across different combinations of graph convolution operators (i.e., GraphSAGE and GAT, here denoted as 'SAGEConv' and 'GATConv', respectively) and underlying graph topologies (i.e., slice-based and encoding-based). Note, as the graph structure is predefined for slice-based topologies, we only evaluate encoding-based methods for different numbers of node neighbours $k$. Every row in the plot depicts the results for a single dataset of the MedMNIST3D initiative. The best-performing setting in terms of graph convolution operator and graph topology is highlighted with a yellow star for each dataset. For comparison reasons, we illustrate the mean performance of the MLP classification head across 3 runs using a dashed horizontal line.} 
\label{fig:block_diagram}
\end{figure}

%% file: images/quantitative_results.tex

\resizebox{\textwidth}{!}{%
\begin{tabular}
{l|
>{\columncolor[HTML]{FFFFFF}}l |
>{\columncolor[HTML]{FFFFFF}}c 
>{\columncolor[HTML]{FFFFFF}}c |
>{\columncolor[HTML]{FFFFFF}}c ||
>{\columncolor[HTML]{EFEFEF}}c }
\toprule
\cellcolor[HTML]{FFFFFF}                                                              & \cellcolor[HTML]{FFFFFF}                                 & \multicolumn{2}{c|}{\cellcolor[HTML]{FFFFFF}\textbf{Metrics}}                                                                & \multicolumn{1}{l||}{\cellcolor[HTML]{FFFFFF}\textbf{Runtime}} & \cellcolor[HTML]{FFFFFF}\textbf{GNN}                                                                         \\ \cline{3-4}
\multirow{-2}{*}{\cellcolor[HTML]{FFFFFF}\textbf{Dataset}}                            & \multirow{-2}{*}{\cellcolor[HTML]{FFFFFF}\textbf{Model}} & \multicolumn{1}{c|}{\cellcolor[HTML]{FFFFFF}\textbf{AUROC $\uparrow$}}  & \textbf{ACC $\uparrow$}                            & \textbf{(min)}                                                & \multicolumn{1}{l}{\cellcolor[HTML]{FFFFFF}\textbf{Configuration}}                                           \\ \midrule
                                                                                      & Baseline                                                 & \multicolumn{1}{c|}{\cellcolor[HTML]{FFFFFF}0.996}                      & 0.907                                              & n/a                                                           & \cellcolor[HTML]{EFEFEF}                                                                                     \\
                                                                                      & DINOv2-MLP                                              & \multicolumn{1}{c|}{\cellcolor[HTML]{FFFFFF}\textbf{0.997} $\pm$ 0.001} & 0.933 $\pm$ 0.002                                  & 3.9                                                           & \cellcolor[HTML]{EFEFEF}                                                                                     \\
\multirow{-3}{*}{\textbf{\begin{tabular}[c]{@{}l@{}}Organ\\ MNIST3D\end{tabular}}}    & \cellcolor[HTML]{EFEFEF}DINOv2-GNN                       & \multicolumn{1}{c|}{\cellcolor[HTML]{EFEFEF}\textbf{0.997} $\pm$ 0.001} & \cellcolor[HTML]{EFEFEF}\textbf{0.943} $\pm$ 0.003 & \cellcolor[HTML]{EFEFEF}\textbf{1.6}                          & \multirow{-3}{*}{\cellcolor[HTML]{EFEFEF}\begin{tabular}[c]{@{}c@{}}SAGEConv\\ manhattan ($L_1$)\\ k=7\end{tabular}} \\ \hline
                                                                                      & Baseline                                                 & \multicolumn{1}{c|}{\cellcolor[HTML]{FFFFFF}0.914}                      & 0.874                                              & n/a                                                           & \cellcolor[HTML]{EFEFEF}                                                                                     \\
                                                                                      & DINOv2-MLP                                              & \multicolumn{1}{c|}{\cellcolor[HTML]{FFFFFF}0.905 $\pm$ 0.001}          & 0.866 $\pm$ 0.003                                  & 3.8                                                           & \cellcolor[HTML]{EFEFEF}                                                                                     \\
\multirow{-3}{*}{\textbf{\begin{tabular}[c]{@{}l@{}}Nodule\\ MNIST3D\end{tabular}}}   & \cellcolor[HTML]{EFEFEF}DINOv2-GNN                       & \multicolumn{1}{c|}{\cellcolor[HTML]{EFEFEF}\textbf{0.918} $\pm$ 0.004} & \cellcolor[HTML]{EFEFEF}\textbf{0.876} $\pm$ 0.004 & \cellcolor[HTML]{EFEFEF}\textbf{1.4}                          & \multirow{-3}{*}{\cellcolor[HTML]{EFEFEF}\begin{tabular}[c]{@{}c@{}}SAGEConv\\ custom\\ k=n/a\end{tabular}}  \\ \hline
                                                                                      & Baseline                                                 & \multicolumn{1}{c|}{\cellcolor[HTML]{FFFFFF}0.750}                      & 0.517                                              & n/a                                                           & \cellcolor[HTML]{EFEFEF}                                                                                     \\
                                                                                      & \multicolumn{1}{c|}{\cellcolor[HTML]{FFFFFF}DINOv2-MLP} & \multicolumn{1}{c|}{\cellcolor[HTML]{FFFFFF}0.812 $\pm$ 0.003}          & \textbf{0.641} $\pm$ 0.004                         & 3.6                                                           & \cellcolor[HTML]{EFEFEF}                                                                                     \\
\multirow{-3}{*}{\textbf{\begin{tabular}[c]{@{}l@{}}Fracture\\ MNIST3D\end{tabular}}} & \multicolumn{1}{c|}{\cellcolor[HTML]{EFEFEF}DINOv2-GNN}  & \multicolumn{1}{c|}{\cellcolor[HTML]{EFEFEF}\textbf{0.819} $\pm$ 0.009} & \cellcolor[HTML]{EFEFEF}0.637 $\pm$ 0.011          & \cellcolor[HTML]{EFEFEF}\textbf{1.5}                          & \multirow{-3}{*}{\cellcolor[HTML]{EFEFEF}\begin{tabular}[c]{@{}c@{}}GATConv\\ chebyshev ($L_\infty$)\\ k=5\end{tabular}}  \\ \hline
                                                                                      & Baseline                                                 & \multicolumn{1}{c|}{\cellcolor[HTML]{FFFFFF}0.839}                      & 0.754                                              & n/a                                                           & \cellcolor[HTML]{EFEFEF}                                                                                     \\
                                                                                      & \multicolumn{1}{c|}{\cellcolor[HTML]{FFFFFF}DINOv2-MLP} & \multicolumn{1}{c|}{\cellcolor[HTML]{FFFFFF}0.926 $\pm$ 0.003}          & 0.870 $\pm$ 0.004                                  & 3.9                                                           & \cellcolor[HTML]{EFEFEF}                                                                                     \\
\multirow{-3}{*}{\textbf{\begin{tabular}[c]{@{}l@{}}Adrenal\\ MNIST3D\end{tabular}}}  & \multicolumn{1}{c|}{\cellcolor[HTML]{EFEFEF}DINOv2-GNN}  & \multicolumn{1}{c|}{\cellcolor[HTML]{EFEFEF}\textbf{0.938} $\pm$ 0.002} & \cellcolor[HTML]{EFEFEF}\textbf{0.883} $\pm$ 0.011 & \cellcolor[HTML]{EFEFEF}\textbf{1.7}                          & \multirow{-3}{*}{\cellcolor[HTML]{EFEFEF}\begin{tabular}[c]{@{}c@{}}GATConv\\ chebyshev\\ k=7\end{tabular}}  \\ \hline
                                                                                      & Baseline                                                 & \multicolumn{1}{c|}{\cellcolor[HTML]{FFFFFF}0.851}                      & 0.795                                              & n/a                                                           & \cellcolor[HTML]{EFEFEF}                                                                                     \\
                                                                                      & \multicolumn{1}{c|}{\cellcolor[HTML]{FFFFFF}DINOv2-MLP} & \multicolumn{1}{c|}{\cellcolor[HTML]{FFFFFF}0.885 $\pm$ 0.001}          & \textbf{0.873} $\pm$ 0.003                         & 4.1                                                           & \cellcolor[HTML]{EFEFEF}                                                                                     \\
\multirow{-3}{*}{\textbf{\begin{tabular}[c]{@{}l@{}}Synapse\\ MNIST3D\end{tabular}}}  & \multicolumn{1}{c|}{\cellcolor[HTML]{EFEFEF}DINOv2-GNN}  & \multicolumn{1}{c|}{\cellcolor[HTML]{EFEFEF}\textbf{0.892} $\pm$ 0.004} & \cellcolor[HTML]{EFEFEF}0.868 $\pm$ 0.004          & \cellcolor[HTML]{EFEFEF}\textbf{1.6}                          & \multirow{-3}{*}{\cellcolor[HTML]{EFEFEF}\begin{tabular}[c]{@{}c@{}}SAGEConv\\ custom\\ k=n/a\end{tabular}}  \\ \hline
                                                                                      & Baseline                                                 & \multicolumn{1}{c|}{\cellcolor[HTML]{FFFFFF}0.930}             & 0.928                                    & n/a                                                           & \cellcolor[HTML]{EFEFEF}                                                                                     \\
                                                                                      & \multicolumn{1}{c|}{\cellcolor[HTML]{FFFFFF}DINOv2-MLP} & \multicolumn{1}{c|}{\cellcolor[HTML]{FFFFFF}0.909 $\pm$ 0.001}          & 0.899 $\pm$ 0.005                                  & 4.2                                                           & \cellcolor[HTML]{EFEFEF}                                                                                     \\
\multirow{-3}{*}{\textbf{\begin{tabular}[c]{@{}l@{}}Vessel\\ MNIST3D\end{tabular}}}   & \multicolumn{1}{c|}{\cellcolor[HTML]{EFEFEF}DINOv2-GNN}  & \multicolumn{1}{c|}{\cellcolor[HTML]{EFEFEF} \textbf{0.918} $\pm$ 0.002}          & \cellcolor[HTML]{EFEFEF} \textbf{0.901} $\pm$ 0.004          & \cellcolor[HTML]{EFEFEF}\textbf{2.0}                          & \multirow{-3}{*}{\cellcolor[HTML]{EFEFEF}\begin{tabular}[c]{@{}c@{}}GATConv\\ manhattan ($L_1$)\\ k=7\end{tabular}}  \\ \bottomrule
\end{tabular}}